\documentclass[10pt,twocolumn,letterpaper]{article}

\usepackage[pagenumbers]{cvpr} 

\usepackage{graphicx}
\usepackage{amsmath}
\usepackage{amssymb}
\usepackage{booktabs}
\usepackage{paralist}

\usepackage[pagebackref,breaklinks,colorlinks]{hyperref}

\usepackage[capitalize]{cleveref}
\crefname{section}{Sec.}{Secs.}
\Crefname{section}{Section}{Sections}
\Crefname{table}{Table}{Tables}
\crefname{table}{Tab.}{Tabs.}

\def\cvprPaperID{*****} 
\def\confName{CVPR}
\def\confYear{2022}

\begin{document}

\title{Document Navigability: A Need for Print-Impaired}

\author{Anukriti Kumar \hspace{10pt} Tanuja Ganu \hspace{10pt} Saikat Guha \\
Microsoft Research, India \\
{\tt\small \{t-anukumar, tanuja.ganu, saikat \}@microsoft.com}}


\maketitle

\begin{abstract}

Printed documents continue to be a challenge for blind, low-vision, and other print-disabled (BLV) individuals. In this paper, we focus on the specific problem of (in-)accessibility of internal references to citations, footnotes, figures, tables and equations. While sighted users can flip to the referenced content and flip back in seconds, linear audio narration that BLV individuals rely on makes following these references extremely hard. We propose a vision based technique to locate the referenced content and extract metadata needed to (in subsequent work) inline a content summary into the audio narration. We apply our technique to citations in scientific documents and find it works well both on born-digital as well as scanned documents. 

\end{abstract}

\section{Introduction}
\label{sec:intro}


Bookshare~\cite{bookshare} and Sugamya Pustakalaya~\cite{sugamyapustakalaya} are libraries of accessible content used by blind, low-vision, and otherwise print-disabled (BLV) individuals. Volunteers for the non-profit scan and digitize books requested by its members. However, it currently takes volunteers close to one month to create accessible books despite the availability of highly-accurate cloud-based OCR services~\cite{google,acs}. This slow pace of making content accessible naturally results in an acute lack of accessible content for BLV individuals. Our goal in this paper is to design tools that empower volunteers to make content accessible an order of magnitude faster.

Making OCR output \emph{navigable} is the primary time-consuming task volunteers perform. This includes extracting chapter, heading and subheading structure, necessary for directly navigating to sections of interest or to skip over content. Automatically detecting heading levels is, in principle, easy using heuristics on font size and weight. A significantly harder problem in making content navigable is resolving internal references such as citations, footnotes, and references to tables, images, and equations in a form that screen-readers can make accessible to BLV individuals.

To illustrate the (in-)accessibility of internal references, consider the following sentence from a research paper: ``[52] presents a breakthrough result''. A sighted reader can effortlessly flip to the references at the back, look up reference~[52], and continue reading within just a few seconds. This effortless lookup is not possible using screen-readers that, in general, do not have a notion of internal references. While internal hyperlinks can be used to approximate the task of jumping to the references, and then navigating back, the interaction disrupts the flow of reading. A better approach is to replace the original sentence with: ``XYZ presents a breakthrough result''.

In order to automatically replace internal references, one must be able to \begin{inparaenum}[i)] 
\item identify internal references in text, 
\item resolve them to the destination content,
\item construct a short readable representation, and
\item replace the original reference with the short representation in a grammatically correct manner.
\end{inparaenum}
In this paper, we focus on solving the first two problems, that of identifying and resolving internal references to citations, tables and figures, and footnotes. In doing so, we discover the lack of tools and datasets for training the necessary models.

Overall, this paper makes three contributions. First, we present our tool for generating ground-truth data on internal references by analysing digital PDFs. Second, we present a dataset for training computer vision models for internal references by applying our tool to several thousand arXiv papers. And finally, we train a vision model on this dataset and apply to a sampling of images of scanned research papers to demonstrate viability of our approach.

\section{Background and Related Work}

PDF has established itself as the de facto standard for fixed-format document exchange and publication but when it comes to reading these documents, a vast majority (75.1 percent) of screen reader users believe that PDF documents are extremely or moderately likely to cause severe accessibility concerns \cite{ref4}. In two earlier research studies, BLV users also reported inaccessible PDFs as a huge barrier in understanding content within a document \cite{ref5},\cite{ref6} as assistive technologies experience significant difficulty trying to interpret the information when the PDF documents do not comply with accessibility standards. Most PDF documents are intrinsically inaccessible, due to the combination of visual layout information and semantic content. 

The PDF Association has also created the Matterhorn Protocol \cite{ref7}, which defines a precise list of 136 test requirements that PDF documents must meet in order to be accessible. This contains detailed guidance for content authors on how to add specifications to their documents and make them compatible with assistive technologies \cite{ref8}. Despite the existence of these thorough guidelines, improving the accessibility of a PDF document remains a challenging research problem due to a combination of limited tools, lack of knowledge and the structure of PDF format itself. Although there are some tools like PAC 3 ~\cite{ref9}, WebAIM’s WAVE ~\cite{ref10} etc. which can identify the problems related to accessibility, there are limited tools that can remediate the issues identified. There are very few tools which are open-source and provide a free version to the user. One of the most commonly used tools for remediation is Adobe Acrobat Pro/DC but its accessibility toolkit is a part of the paid version and hence, very few people have access to it. 

Most research articles that are delivered today exist in PDF format, and a recent study showed that only 2.4\% of them are accessible to people with disabilities \cite{ref11}. Recent research has focused on addressing some important aspects of paper accessibility, such as how screen readers should comprehend and read mathematical equations \cite{ref12, ref13, ref14, ref15}, generate figure captions automatically \cite{ref16, ref17} and explain graphs and charts \cite{ref18, ref19, ref20}. In the recent work, a lot of traction is given upon improving accessibility of various types of media content such as images \cite{ref21, ref22}, videos \cite{ref23} and automatic classification of the content of figures \cite{ref24}. Other work focused on improving automatic text and layout detection in scanned documents \cite{ref25} as well as table content extraction \cite{ref26, ref27} within them. To date, not much work has been done for introducing navigability within a document which enables skimming and scanning \cite{ref28}, currently under-supported by PDF documents. Hence, our work focuses on taking the first step towards automatically replacing the internal references and hence, improving the reading experience of a wide range of users.

\section{Methodology}
\label{method-section}
\begin{figure}[h]

  \centering
  \includegraphics[width=0.75\linewidth]{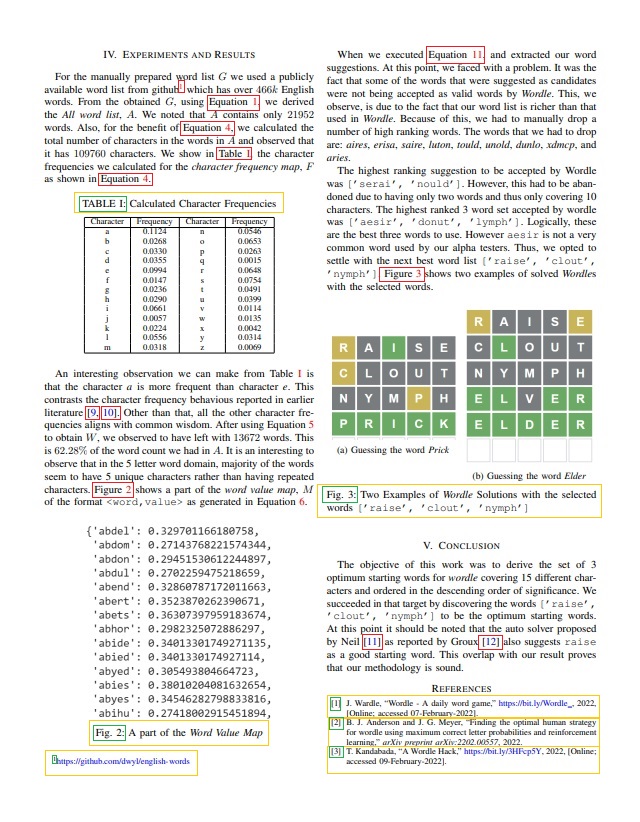}

  \caption{A document page decomposed into regions of interest (yellow box), explicit keys (green box) and implicit keys (red box) for tables, figures, references and footnotes. 
}
  \label{fig:methodology}
  \vspace{-5pt}
\end{figure}

This section talks about the tool we created to generate ground-truth data on internal references by analyzing digital PDFs. The process for building this tool is divided into 3 phases: First, we segment the document into regions of interest. Second, we identify explicit internal reference keys for these regions. And finally, we infer implicit keys associated with references using the hyperlink text. These three key elements are depicted in Fig. \ref{fig:methodology}.

\begin{figure}[t]
  \centering
  \includegraphics[width=0.75\linewidth]{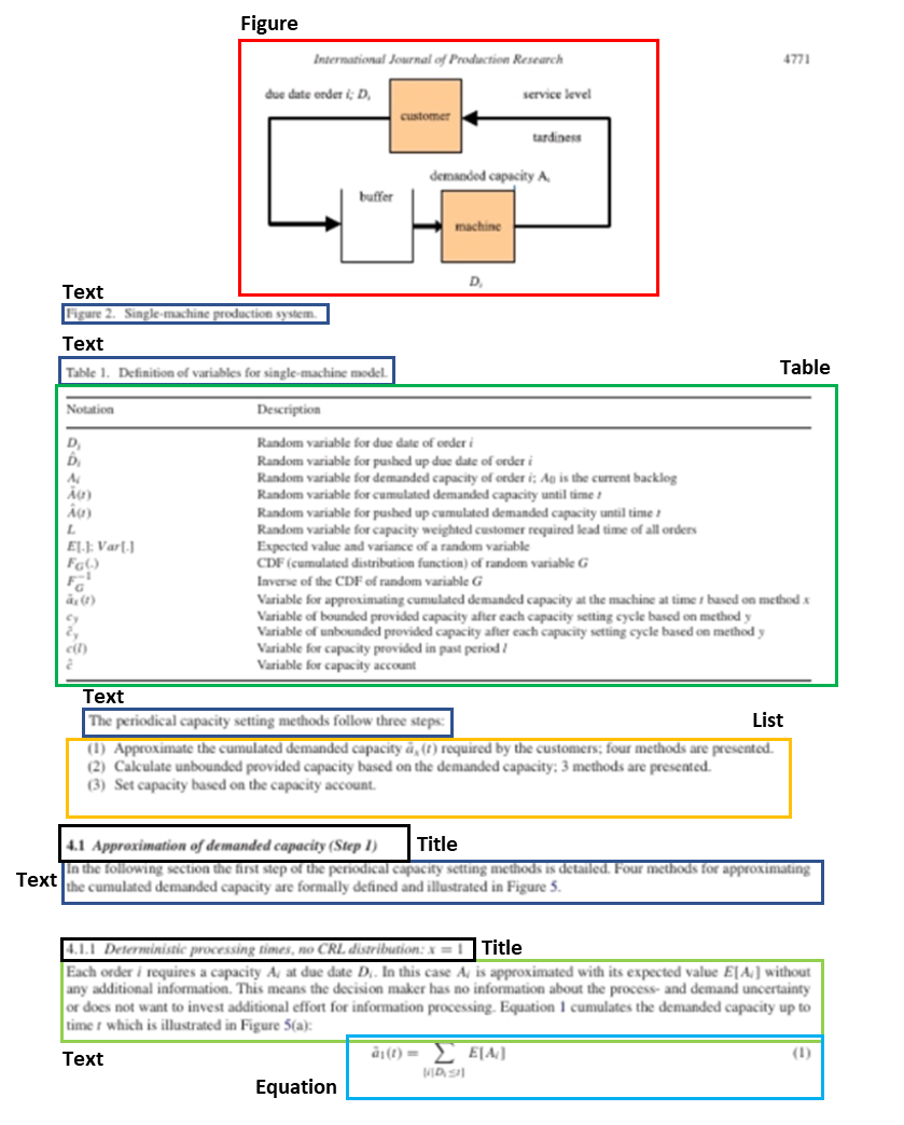}

  \caption{Result from document segmentation model: A document page is segmented into six categories: Text, Title, List, Table, Figure and Equation.). 
}
  \label{fig:segmentation model}
  \vspace{-10pt}
\end{figure}

\textbf{Regions.} A research paper can be decomposed into the following elements: headings, text blocks, figures, tables, lists and equations. For detecting internal references within these elements, we need to first identify the regions containing internal references and we pose this problem as an instance segmentation task. We use a pre-trained model based on Mask R-CNN architecture from Detectron2 model zoo to decompose a document into five categories: title, text block, list, figure and table. The model is based on the ResNet50 feature pyramid network (FPN) base config and is trained on the Publaynet dataset for document layout analysis. Similarly, we use another instance segmentation model of the same base config to detect equations but the only difference is that it has been trained on MFD (Mathematical Formula Detection) dataset with output layer modified to predict only one category: equation. The result obtained from these models is depicted in Fig. \ref{fig:segmentation model}.

\textbf{Explicit Keys.} After extracting the regions of interest, our focus is to extract explicit internal reference keys for each of these regions. This involves identification of reference items, table captions, figure captions, equation numbers and footnote markers. However, there is a dearth of tools and data which makes it difficult to extract them directly. Hence, we use different heuristics for extracting such elements. Since we already know certain standardized keywords which mark the beginning of these elements, we make use of OCR to extract text within the regions and hence, create appropriate annotations for the internal references. These keywords are extracted as explicit keys, used for resolving reference items later. For instance, a table or figure is generally surrounded by its caption which can be segmented as a text block from the document segmentation model. Now, doing OCR on these text blocks and a simple keyword matching would give us the appropriate caption. Similarly, superscript detection would provide us with the footnote markers. Using the annotations obtained from this process, we create a structured representation for all reference items with fields for bounding box coordinates as well as text contained within them.
\vspace{-10pt}

\textbf{Implicit Keys.} Within a scientific document, there can be multiple hyperlinks associated with the same internal reference. Since the objective of our work is to create a dataset for resolving internal references with their corresponding in-text citations and vice-versa, we need to obtain all hyperlink texts which would act as implicit keys for internal references. Firstly, we incorporate document’s metadata using a python module named fitz \cite{fitz} from an open source project PyMuPDF. This way we collect information regarding the source location as well as the target point location for all types of embedded links. For each target point location, we find the nearest reference item from all the items extracted earlier. Also, we extract all implicit keys by applying OCR at all hyperlink locations obtained from fitz. Finally, a ground-truth dataset is created by combining source keys (implicit and explicit keys) with the annotated reference bounding box and associated text.


\section{Experiments and Result}
This section focuses on applying our tool to generate ground-truth annotated data for bibliographic reference items present in scientific documents. Further, we would also demonstrate the viability of our approach by training a vision model on this data for bibliographic reference item detection and applying it to both born-digital as well as scanned documents. The entire pipeline is described in the following sub-sections.

\subsection{Data}
\label{data}
The first step is to construct a dataset of research papers by sampling PDFs published in the years of 2016-2021 from arXiv dataset published on Kaggle \cite{kaggle}, stratified across subjects ranging from physics, computer science, mathematics, physics, statistics, quantitative biology, economics and others. We sampled papers from each field of study, with Physics on the minimum end and Economics and Computer Science on the maximum end. The resulting dataset consisted of around 22,081 papers. It is used for generating ground-truth labels for bibliographic reference items and also, for training the vision model to detect bibliographic reference items.

\subsection{Ground-Truth Labels}
This section talks about generating ground-truth labels for bibliographic reference items by applying our tool to research paper dataset as mentioned in section \ref{data}.

To identify the regions of interest for bibliographic references, we utilized the document segmentation model as described in section \ref{method-section} to detect two classes of elements - Title and List blocks within a research paper. Then, we extracted text within the title blocks using OCR to check for references section. Further, a keyword search based heuristic is used to identify phrases that are likely to appear as a title for the references section. Once we detected the section, we segmented out list blocks within it and this way, we extracted 88,786 list images containing references.

Next, we segmented reference items out of the list images by exploiting some natural properties of reference items, such as the use of certain characters that are used to mark the beginning of a reference item. We have also assumed that they are written in a consistent manner as is required by most academic venues. But since these items can span through multiple lines, the detection becomes difficult. Hence, we used a heuristics based approach based on identifying these start characters from references using OCR and further, labeling the region between them as a reference item. Further, these start characters are extracted as an explicit key for the reference.


To extract implicit keys, we obtained embedded links information using fitz module. Then, we filtered out only those in-text hyperlinks which were pointing to pages beyond reference section. For each hyperlink, we detected its corresponding reference item by finding an annotated bounding box nearest to hyperlink's target point location. Also, we obtained implicit keys by extracting hyperlink text from all the locations pointing to a bibliographic reference. Finally, we combined all these key elements to generate annotated ground-truth data for bibliographic reference items.

\subsection{Model}
To generalize the process of detecting all types of references and linking them back to their in-text citations, we trained a vision based model for extraction of bibliographic reference items along with their source keys. We used the base config from the Faster-RCNN model present in the model zoo of the Detectron2 library. It is pre-trained on COCO dataset for object detection task and is based on the ResNet101 feature pyramid network (FPN) base config. This backbone network is used to extract features from an image followed by a Region Proposal Network for proposing region proposals and a box head for tightening the bounding box. 
To train this model on our dataset, we first divided our ground-truth dataset into two parts as train and validation set with 0.85 as the split ratio. Out of 88,786 list images data and their annotations, 75,468 were used for training and the remaining 13,318 were utilized for evaluation purposes. The model was trained for 20k iterations.


\begin{table}[h]
  \centering
  \setlength{\tabcolsep}{4pt}
  \begin{tabular}{@{}c|cccccc@{}}
    \toprule
    \footnotesize Model & \footnotesize AP & \footnotesize AP@$.50$ & \footnotesize AP@$.75$ & \footnotesize AP$_{m}$ & \footnotesize AP$_l$ \\
    \midrule
    \footnotesize Reference segmentation & \footnotesize 81.70 & \footnotesize 86.98 & \footnotesize 84.47 & \footnotesize 74.68 & \footnotesize 84.72 \\ 
    \bottomrule
  \end{tabular}
  \caption{Average precision values at different IoU thresholds}
  \label{tab:apscore}
  \vspace{-5pt}
\end{table}

\subsection{Results}
We inferred the results by testing our model on validation dataset using hold-out cross validation technique. Usually, object detection models are evaluated following the COCO standards of evaluation. Hence, mAP (mean average precision) is an important metric considered for evaluating the performance of the model. The average precision values at different IoU thresholds are present in Table \ref{tab:apscore}.

\begin{figure}[t]
  \centering
  \includegraphics[width=0.7\linewidth]{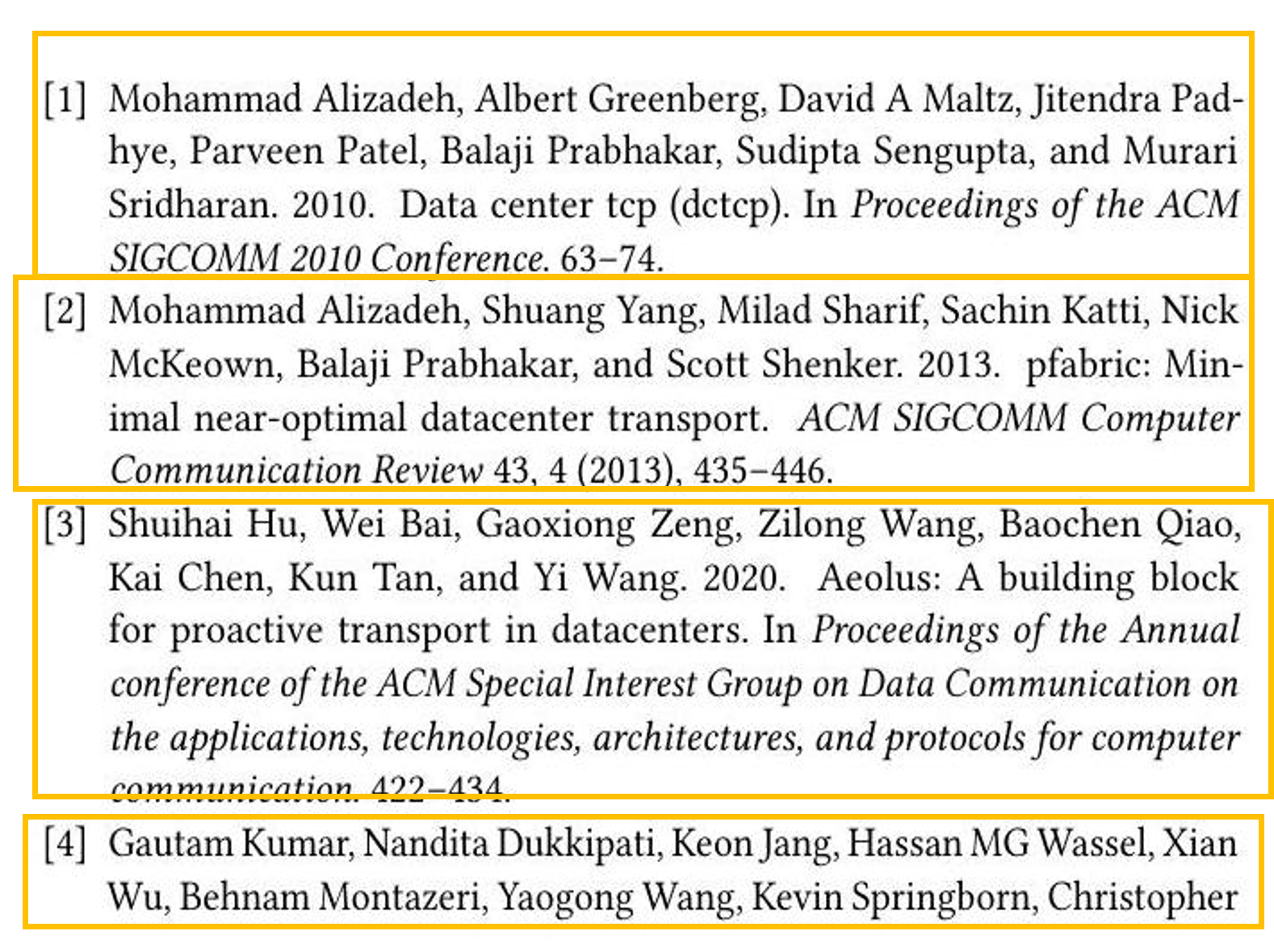}

  \caption{Output for research paper image from born-digital PDF
}
  \label{fig:sample output1}
\end{figure}

\begin{figure}[t]
  \centering
  \includegraphics[width=0.7\linewidth]{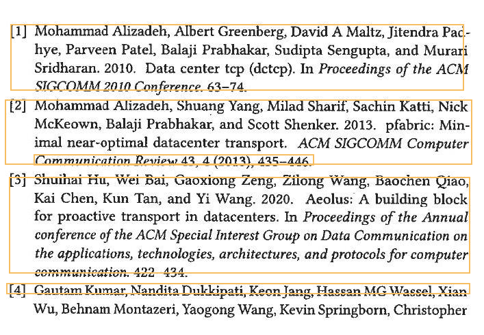}

  \caption{Output for scanned research paper image 
}
  \label{fig:sample output2}
\end{figure}

We also validated our approach by applying it in two different settings - 1) images from born-digital PDFs and 2) images from a flatbed scanner. The output is shown in Fig. \ref{fig:sample output1} and Fig. \ref{fig:sample output2} respectively. In both cases, we were able to extract reference items and their source keys successfully. This depicts the viability of our approach and hence, it can be used to identify and resolve other types of internal references as well.

\section{Discussion}
From Fig. \ref{fig:sample output1} and Fig. \ref{fig:sample output2}, we observe that the performance of our model is significantly better on images from born-digital PDF than a real-world scanned image. Hence, one of the future directions is to make our model more robust to data distribution shifts through data augmentation. 

For some research paper images, we also observed false negatives for reference items continuing from the previous page. We plan to handle this case by adding more data annotations of this type in future. 

Also, we found that exploiting some natural properties of reference items, such as the use of certain keywords in the beginning of a reference item, makes heuristic approaches very effective for our task. However, in future, we plan to use  a considerably expanded set of keywords to make our approach effective against more kinds of papers.

\section{Conclusion}

In this paper we presented our ongoing work in making published documents accessible to blind, low-vision, and print-disabled individuals. We specifically focused on the problem of poor accessibility of internal references like citations, footnotes, table and figure references. We presented a vision based technique to extract metadata needed to make these internal references accessible. We successfully applied our technique to extract requisite metadata from the bibliography section. We continue to work on summarizing the referenced content and inlining into the audio narration to make internal references fully accessible to print-impaired individuals.   

{\small
\bibliographystyle{ieee_fullname}
\bibliography{egbib}
}

\end{document}